\documentclass[10pt,twocolumn,letterpaper]{article}

\usepackage{cvpr}              
\definecolor{cvprblue}{rgb}{0.21,0.49,0.74}
\usepackage[pagebackref,breaklinks,colorlinks,allcolors=cvprblue]{hyperref}

\def\paperID{} 
\def\confName{CVPR}
\def\confYear{2026}

\title{\textit{Dual Anchors, Do It Better}: Hierarchical Group Merging \\
for Zero-Shot Anomaly Detection
}

\usepackage{marvosym}
\author{
\textbf{Jimin Roh$^{1,2*}$} \qquad
\textbf{DongKyu Kim$^{1,3*}$} \qquad
\textbf{Suk-Ju Kang\textsuperscript{1,\Letter}} \\[0.5em]
{$^{1}$Sogang University,} 
{$^{2}$LG Electronics,} 
{$^{3}$NAVER Cloud} \\
{\ttfamily\small
shwlals96@gmail.com \quad
kyu15911@gmail.com \quad
sjkang@sogang.ac.kr
} \\
{\small
$^{*}$ Equal contribution \hspace{1.5em}
\Letter \hspace{0.5mm}Corresponding author
}
}

\usepackage{kotex}
\usepackage{array}
\usepackage{booktabs}
\usepackage{multirow}
\usepackage{makecell}
\usepackage{graphicx}
\usepackage{caption}

\usepackage[table]{xcolor} 
\usepackage{colortbl}     
\definecolor{lightgreen}{RGB}{233,245,233}
\definecolor{darkred}{RGB}{180,0,0}

\usepackage{pifont}
\usepackage[accsupp]{axessibility}

\begin{document}
\twocolumn[{
\renewcommand\twocolumn[1][]{#1}
\maketitle
\begin{center}
    \setlength{\abovecaptionskip}{-2pt} 
    \setlength{\belowcaptionskip}{-4pt}
    \centering
    \resizebox{\textwidth}{!}{%
    \includegraphics[height=1.0\textwidth]{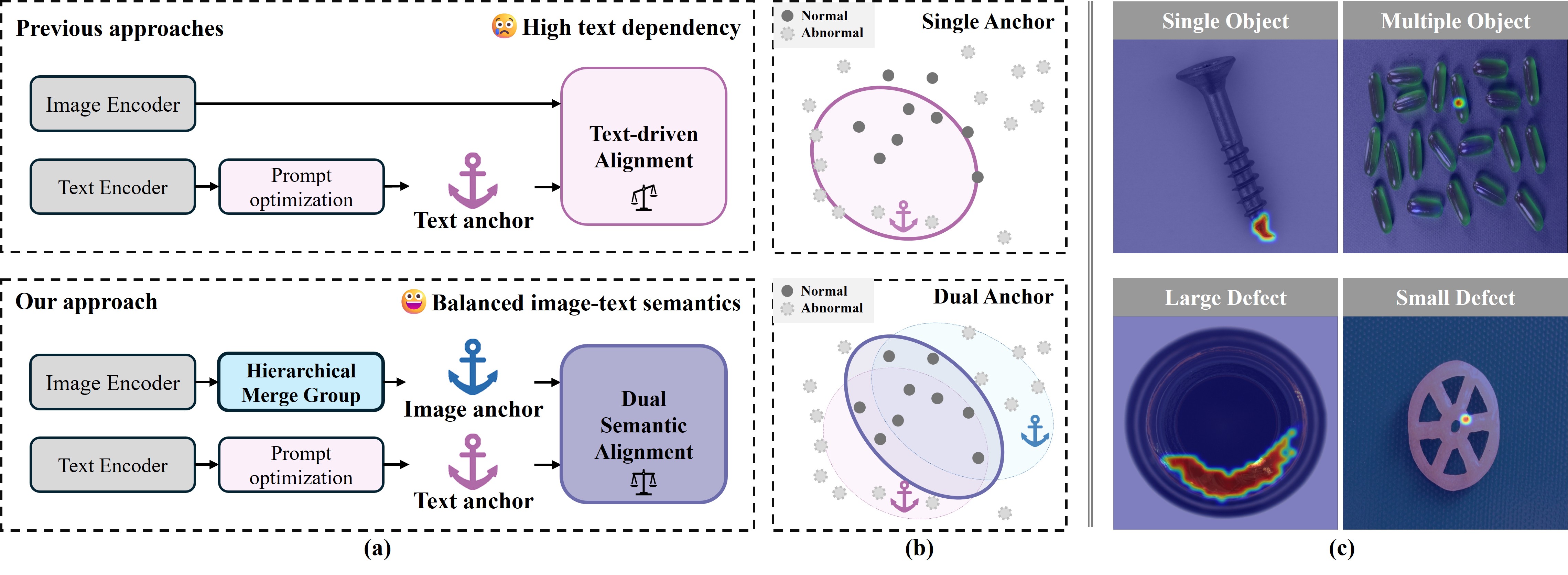}
    }
    \captionof{figure}{
    \textbf{Overview of the proposed Dual-Anchor  for Zero-Shot Anomaly Detection.}
Previous methods perform \textit{text-driven alignment}, anchoring semantics only on text prompts, which causes excessive text dependency and weak visual grounding. Our approach introduces a \textit{Hierarchical Group Merging} that constructs \textit{image semantic anchors} and fuses them with text anchors, forming a \textit{dual-anchor paradigm} that achieves balanced alignment and clearer semantic separation.
The proposed framework ensures consistent anomaly localization across diverse object quantities and defect scales, demonstrating robust generalization and stable anomaly reasoning.
    }
    \label{fig1}
\end{center}
}]

\begin{abstract}
Zero-shot anomaly detection (ZSAD) aims to identify anomalies in unseen domains, a setting that is particularly critical for industrial and medical applications where domain shifts are prevalent. However, most CLIP-based ZSAD methods anchor semantics solely on the text modality, making performance highly sensitive to prompt design and leading to weak visual grounding. To mitigate these limitations, we propose a \textbf{Dual-Anchor framework} that complements conventional text anchors with hierarchical image anchors constructed via a top-down grouping mechanism. This mechanism progressively aggregates local-to-global image features to form normal and abnormal group tokens, which serve as image anchors and act as gating signals in a Group-Gated Token Refiner to enhance the global representation. The refined image anchors are then fused with text prompts to construct dynamic state prompts. By jointly reinforcing visual and textual semantics, our framework stabilizes image-text alignment, reduces prompt dependency, and achieves strong generalization across 8 industrial and 6 medical benchmarks.
\end{abstract}
\vspace{-4mm}
\section{Introduction}
\label{sec:intro}

Anomaly detection has become increasingly crucial in high-reliability domains such as industrial inspection and medical diagnostics. In particular, in scenarios characterized by high annotation costs and significant domain shifts, there is a growing demand for approaches that can achieve robust generalization without relying on pre-collected normal or anomalous data. To address this challenge, Zero-Shot Anomaly Detection (ZSAD), empowered by Vision-Language Models such as CLIP~\citep{CLIP}, has recently emerged as a promising research paradigm. In ZSAD, a model is trained on auxiliary data and then deployed to unseen target domains without any target-domain labels. By leveraging large-scale pre-trained image-text representations, ZSAD enables anomaly detection in unseen target domain and domain-agnostic manner.

Most ZSAD approaches leverage the powerful image-text alignment capability of CLIP, which is trained on large-scale image-text pairs. Early studies~\citep{WinCLIP, deng2023bootstrap} employed manually crafted prompts that explicitly encode semantic cues describing normal and anomalous states, thereby directly utilizing the representational power of CLIP while exploiting multi-layer visual features extracted from the image encoder. 
Subsequent research introduced the concept of learnable prompts to alleviate the reliance on manual design and improve adaptability. These methods learn context-aware prompts that dynamically adjust to visual content by incorporating global or instance-level cues from images, resulting in enhanced generalization across diverse domains~\cite{AdaCLIP, AnomalyCLIP, Bayes-PFL}. 
More recently, several studies have attempted to embed anomalous cues derived from the image itself into the prompt representation, further enriching the expressiveness and scalability of text-guided anomaly detection frameworks.

However, existing studies still suffer from several limitations. First, as illustrated in Fig.~\ref{fig1}(a, top), most methods place the semantic anchors for normal and anomalous concepts primarily on the text prompt side. This design makes the overall performance highly sensitive to prompt construction and tuning, resulting in excessive dependence on textual representations. Consequently, as shown in Fig.~\ref{fig1}(b, top), the decision boundary in the unseen domain fails to encompass all anomalous images. Second, many existing approaches do not fully exploit the structural advantage of image encoders, which inherently capture visual features from local to global levels. Instead, they often treat features extracted from different layers as independent, thereby overlooking the hierarchical integration of semantic information essential for understanding normal and anomalous patterns.

Therefore, as illustrated in Fig.~\ref{fig1}(a, bottom), we propose a \textbf{Dual-anchor framework} that hierarchically merges the features extracted from the image encoder to construct image anchors. Specifically, the proposed method employs a top-down \textit{Hierarchical Group Merging} strategy, which progressively aggregates visual features from local to global levels, thereby enhancing the semantic representations of normal and anomalous concepts in a hierarchical manner. 
During this process, semantically distinct normal and abnormal group tokens are formed and utilized as image anchors. Subsequently, a \textit{Group-Gated Token Refiner} is introduced, which leverages a gating mechanism on the group tokens to strengthen the global image representation. 
As shown in Fig.~\ref{fig1}(b, bottom), unlike previous approaches that rely on a single anchor, the proposed dual-anchor framework mitigates bias toward one anchor and enables a more balanced decision boundary in unseen domains. 
Lastly, the group tokens are integrated with text prompts to construct \textit{Dynamic State Prompts} for normal and abnormal states, thereby reducing the dependency and manual cost associated with prompt engineering.

Finally, we fairly evaluate our dual-anchor framework across eight industrial domains and six medical domains to demonstrate its effectiveness. Our main contributions are as follows:
\begin{itemize}
    \item \textbf{Dual-Anchor Framework.} 
    We propose a novel dual-anchor paradigm that constructs normal and abnormal visual anchors, mitigating bias toward text prompts and enabling balanced decision boundaries in unseen domains.

    \item \textbf{Hierarchical Group Merging.} 
    A top-down merging strategy that progressively aggregates local-to-global visual features, forming semantically enriched image anchors with hierarchical anomaly representations.

    \item \textbf{Dynamic State Prompt.} 
    Group tokens are integrated with text prompts to generate adaptive state-aware prompts, reducing manual design costs and enhancing image-text alignment stability.
\end{itemize}

\section{Related Works}
\subsection{Zero-Shot Anomaly Detection}
\begin{figure*}[t]
\centering
\includegraphics[width=\textwidth]{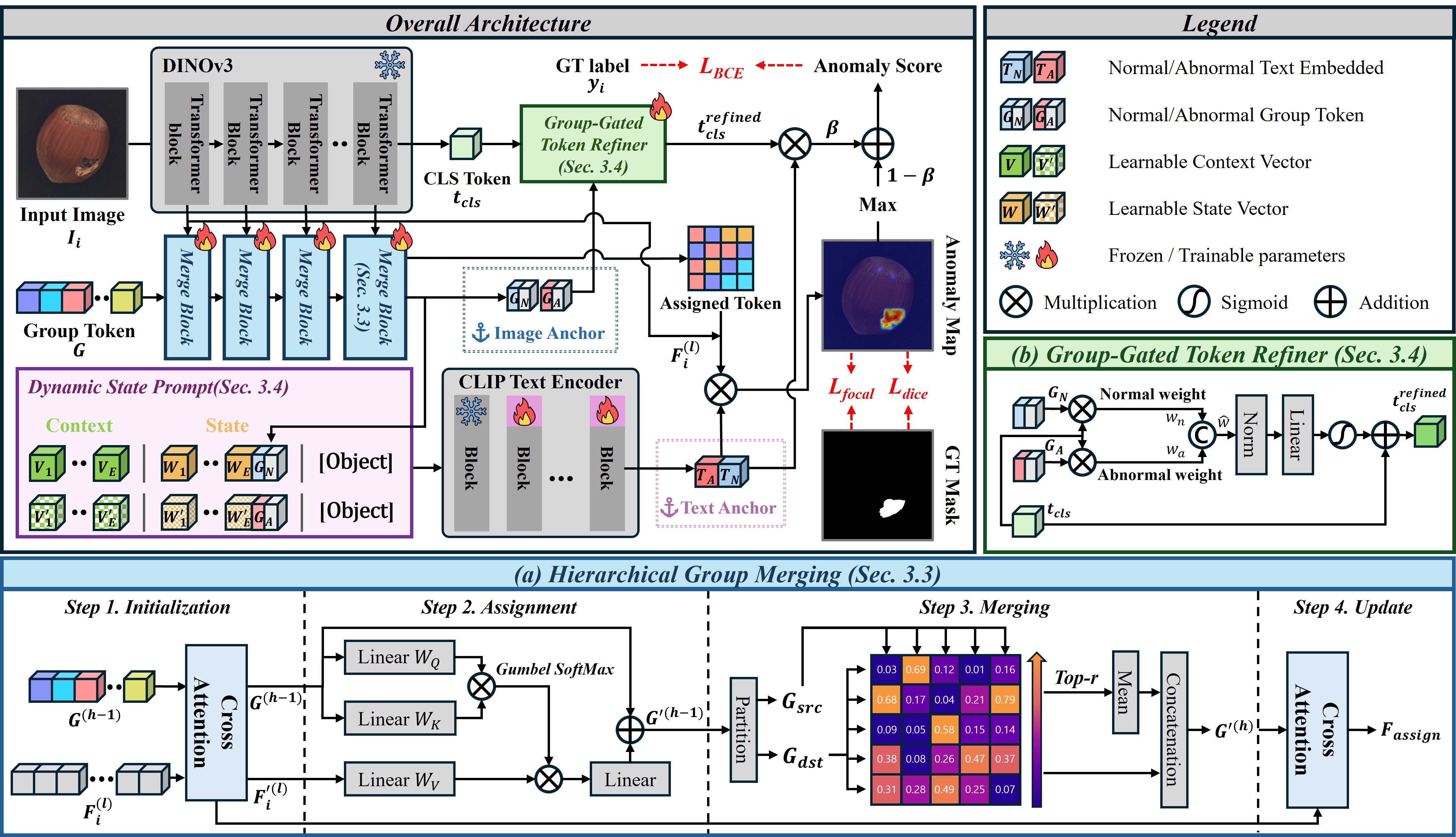}
\caption{
\textbf{Overall Architecture of our framework.}
Our framework adopt frozen DINOv3 backbone (Sec. 3.2) to extract multi-scale patch tokens as visual representations.
Subsequently, the \textit{Hierarchical Group Merging} module aggregates these features with learnable group tokens to construct \textit{image anchors} (Sec. 3.3).
The generated image anchors are then utilized by the \textit{Group-Gated Token Refiner} and the \textit{Dynamic State Prompt} (Sec. 3.4) to refine the global representations and establish balanced dual-anchor alignment between image and text semantics.
}
\label{fig2}
\end{figure*}

Zero-Shot Anomaly Detection (ZSAD) aims to distinguish between normal and abnormal samples that have not been observed during training. Most recent approaches leverage the large-scale image–text alignment capability of Vision–Language Models (VLMs), such as CLIP~\cite{CLIP}, by measuring the semantic similarity between an input image and text prompts that define the concepts of normality and abnormality. This paradigm has attracted growing attention in industrial and medical domains, as it enables anomaly recognition without the need for defective training data.

Early approaches relied on hand-crafted prompts to directly compare features extracted from CLIP’s image and text encoders. However, these methods were highly sensitive to the phrasing of prompts and often exhibited unstable results due to linguistic biases~\cite{WinCLIP, CLIP-AD, APRIL-GAN}. To address this limitation, subsequent studies introduced learnable prompts to reduce language bias~\cite{AnomalyCLIP, AdaCLIP} or modeled the prompt space as a probabilistic distribution to enable diverse prompt sampling and improve generalization~\cite{Bayes-PFL}.

Accordingly, recent works have attempted to alleviate textual dependency by introducing visual-context prompts~\cite{VCP-CLIP}, where image-conditioned cues are incorporated to enrich textual representations. While such approaches mitigate linguistic bias to some extent, they still rely primarily on text-driven alignment and lack structured visual reasoning that captures hierarchical semantics.

Despite these advances, most methods still depend heavily on textual supervision. Thus, robust zero-shot anomaly detection requires not only cross-modal alignment with textual semantics but also image-centric representations, in which visual structures play a central role in semantic reasoning.

\subsection{Hierarchical Grouping}

GroupViT~\cite{GroupViT} introduces an explicit grouping mechanism into Vision Transformers, where patch-level tokens are progressively merged via cross-attention with learnable group tokens to form hierarchical visual structures. Later studies, such as TokenLearner~\cite{Tokenlearner} and ToMe~\cite{ToMe}, further explored adaptive token aggregation to balance semantic expressiveness and computational efficiency. HGFormer~\cite{HGFormer} extends this concept by grouping features across multiple transformer layers to integrate local information and global context within a unified hierarchical framework, explicitly modeling semantic dependencies to enhance spatial and contextual coherence. 

These grouping-based approaches move beyond simple feature aggregation and instead organize representations to capture the intrinsic hierarchy of visual data.

In this work, we utilize hierarchical grouping to organize patch-level representations into semantically coherent groups, focusing on leveraging such structured visual hierarchies to facilitate zero-shot anomaly reasoning without requiring explicit supervision.

\label{sec:related}

\section{Method}
\label{sec:method}

Our framework consists of three main modules: 1) \textbf{\textit{Hierarchical Group Merging}} that generates normal and anomalous group tokens as image anchors, 2) \textbf{\textit{Group-Gated Token Refiner}} that leverages these group tokens to refine and coordinate the global image representation, and 3) \textbf{\textit{Dynamic State Prompt}} that constructs adaptive prompts conditioned on the image state. Collectively, these modules are integrated to achieve robust image-text alignment performance in ZSAD.

\subsection{Problem Setting}
Our task follows the standard ZSAD setting, where the model is trained on an auxiliary dataset and evaluated on unseen target domains. 
Formally, the auxiliary training dataset is defined as $D_{\text{train}} = \{(I_i, y_i)\}_{i=1}^{N_{train}}$, where \( I_i \in \mathbb{R}^{H \times W \times 3} \) and \( y_i \in \{0 \text{ (normal)}, 1 \text{ (abnormal)}\} \). During evaluation, we consider a target test, which contains samples from unseen domains or categories that do not overlap with $D_{\text{train}}$.
\subsection{Feature Extraction.}
Given an input image $I_i$, we extract patch tokens $F_i^{(l)} \in \mathbb{R}^{M \times C}$ from the $l$-th layer of the backbone network, where $M$ denotes the number of patches and $C$ the feature dimension. We adopt DINOv3~\citep{simeoni2025dinov3} as our backbone. While DINOv3 is renowned for its exceptional image representation capability, it exhibits relatively weaker text alignment compared to CLIP~\citep{CLIP}. We acknowledge this trade-off between visual representation power and text-language alignment, and therefore prioritize high quality image features that are more suitable for anomaly localization and representation learning.
\subsection{Hierarchical Group Merging}
Figure~\ref{fig2} (a) illustrates the detailed architecture of a single \textit{Merge Block}. Each consists of four sequential stages designed to progressively refine group representations across layers:
\begin{enumerate}[leftmargin=1.5em, itemsep=0.2em]
    \item \textbf{Initialization:} Initializing the grouping process with previous-layer group representations.
    \item \textbf{Assignment:} Assigning current patch tokens to the corresponding groups based on feature similarity.
    \item \textbf{Merging:} Merging the assigned groups to form higher-level group representations for the next layer.
    \item \textbf{Update:} Update: Re-assigning merged group tokens for inter-layer propagation.
\end{enumerate}
\paragraph{Initialization.}
In the initialization stage, a cross-attention layer is applied between the previous group tokens and the current patch tokens to inherit grouping cues from the preceding layer. 
This design allows hierarchical information captured by the backbone to be progressively accumulated, facilitating efficient grouping without requiring additional sequential feature extraction:
\begin{equation}
F_{i}'^{(l)} = \mathrm{CrossAttn}\!\left(F_{i}^{(l)},\, \mathbf{G}'^{(h)}\right),
\end{equation}
where $\mathbf{G}'^{(h)}\in \mathbb{R}^{K \times C}$ denotes the learnable group tokens at the $h$-th stage of the grouping process.
\paragraph{Assignment.}After obtaining the initialized tokens $F_{i}'^{(l)}\in \mathbb{R}^{M \times C}$, these tokens are assigned to the current groups through a learnable Gumbel Softmax-based assignment~\citep{jang2016categorical, maddison2016concrete}.
This process enables discrete yet differentiable group selection, allowing each patch token to be stochastically associated with one of the candidate groups during training. The Gumbel Softmax function provides a continuous relaxation of categorical sampling, defined as:
\begin{equation}
\pi_i = 
\frac{\exp\!\left((\log \alpha_i + g_i)/\tau\right)}
{\sum_{j=1}^{K} \exp\!\left((\log \alpha_j + g_j)/\tau\right)}, 
\quad g_i \sim \mathrm{Gumbel}(0,1),
\end{equation}
where $\alpha_i$ denotes the unnormalized attention logits, $\tau$ is the temperature controlling the discreteness, and $g_i$ is Gumbel noise sampled from $\mathrm{Gumbel}(0,1)$. By incorporating the Gumbel-Softmax mechanism into cross-attention, the model approximates \textit{hard group assignments} while remaining fully differentiable. This enables stable end-to-end optimization and enhances the semantic discreteness and interpretability of the grouping process.
\begin{equation}
\begin{aligned}
\boldsymbol{\Pi} =
\mathrm{Softmax}\!\left(
\frac{\mathbf{F}_{i}'^{(l)}W_Q(\mathbf{G}^{(h)}W_K)^{\!\top} + \boldsymbol{\Gamma}}{\tau}
\right)
\end{aligned}
\end{equation}
\begin{equation}
\mathbf{G}'^{(h)} =
\mathbf{G}^{(h)} + \mathrm{Linear}\!\big(\boldsymbol{\Pi}(\mathbf{G}^{(h)}W_V)\big)
\end{equation}
where $\boldsymbol{\Gamma}$ denotes the Gumbel noise sampled from $\mathrm{Gumbel}(0,1)$ used for stochastic and differentiable group assignment.
\paragraph{Merging.}
To efficiently merge the group tokens, we adopt a bipartite soft matching strategy~\citep{ToMe}, 
which performs parameter-free token reduction based on pairwise similarity. 
Given the group tokens $\mathbf{G}'^{(h)}\in \mathbb{R}^{K \times C}$, we partition them into a source set 
$\mathbf{G}'_{\mathrm{src}} \in \mathbb{R}^{(K/2) \times C}$ and a destination set $\mathbf{G}'_{\mathrm{dst}} \in \mathbb{R}^{(K/2) \times C}$. 
We then compute the pairwise similarity matrix between $\mathbf{G}'_{\mathrm{src}}$ and 
$\mathbf{G}'_{\mathrm{dst}}$ as:
\begin{equation}
\mathbf{S} = \cos\!\left( \mathbf{G}'_{\mathrm{src}},\, \mathbf{G}'_{\mathrm{dst}} \right),
\end{equation}
where $\cos(\cdot, \cdot)$ denotes the cosine similarity.

Next, for each token $\mathbf{g}_u \in \mathbf{G}'_{\mathrm{src}}$, 
we select the token $\mathbf{g}_v \in \mathbf{G}'_{\mathrm{dst}}$ that maximizes the similarity:
\begin{equation}
v'(u) = \arg\max_{v} \mathbf{S}_{u,v}.
\end{equation}
From the resulting pairs, we keep the top-$r$ matches according to the similarity scores 
to maintain gradual token reduction.

Each matched pair is then merged by averaging their features:
\begin{equation}
\tilde{\mathbf{g}}_{u} = \frac{1}{2}\left( \mathbf{g}_u + \mathbf{g}_{v'(u)} \right).
\end{equation}

Finally, the merged tokens replace the original pairs, and the resulting token set is concatenated 
for further processing:
\begin{equation}
\mathbf{G}'^{(h+1)} = \mathrm{Concat}\!\left( \tilde{\mathbf{G}}',\, \mathbf{G}'_{\mathrm{remain}} \right),
\end{equation}
where $\mathbf{G}'_{\mathrm{remain}}$ denotes the tokens not merged in this process. 
This approach avoids iterative clustering, enables efficient parallel computation, 
and ensures that most tokens remain unmerged, preserving representational fidelity throughout the network.

\paragraph{Update.}
The update stage is applied only in the final \textit{Merge Block}, where the group tokens encapsulating hierarchical semantics are assigned back to the patch tokens to produce the final assigned representation $F_{assign}$.
\begin{figure}[t]
\centering
\includegraphics[width=0.9\columnwidth]{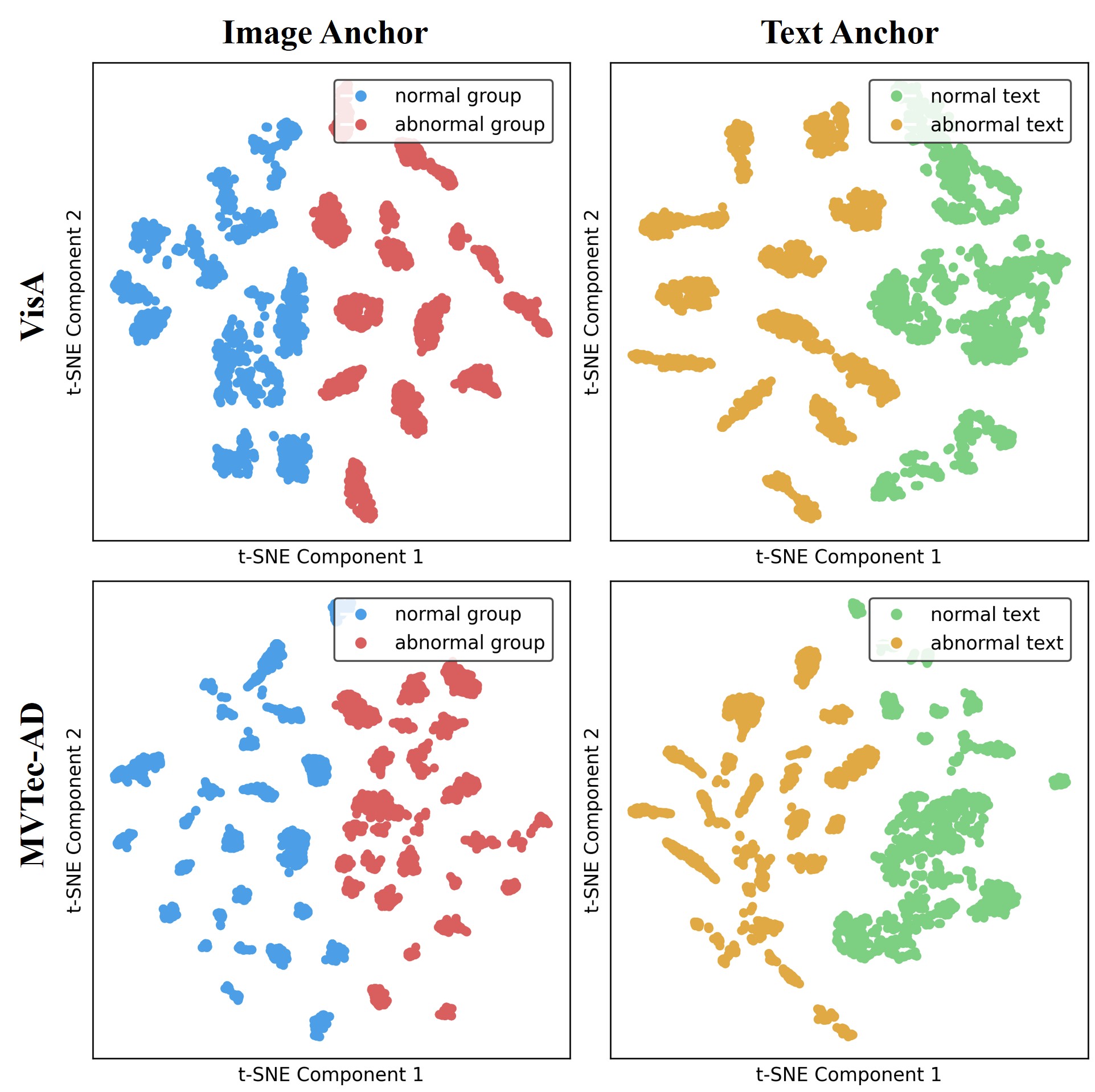}
\caption{
\textbf{T-SNE visualization of image and text anchors on VisA and MVTec-AD datasets.}
}
\label{fig3}
\vspace{-4mm}
\end{figure}
\paragraph{Image Anchor.} 
 Through this top-down hierarchical merging process, information from different levels is progressively integrated, and the model ultimately produces the normal and anomalous group tokens $G_N$, $G_A$
along with the corresponding assigned tokens $A$. This process can be formulated as:
\begin{equation}
\begin{aligned}
\{ \mathbf{G}_N,\, \mathbf{G}_A \},\, F_{assign}
&= \sum_{h=1}^{H} \textit{MergeBlock}\!\left(F^{(h)},\, \mathbf{G}'^{(h)}\right)
\end{aligned}
\end{equation}
Here, $\mathbf{G}_N$ and $\mathbf{G}_A$ denote the first and second group tokens obtained from the final grouping stage, respectively. 
Figure~\ref{fig3} presents the t-SNE visualization of our dual-anchor representation in an unseen domain. 
As shown, the image anchor exhibits a clear separation between normal and abnormal groups, 
demonstrating that our method effectively establishes distinct visual semantics for each state. 
Moreover, this result highlights that our approach mitigates the dependence on text anchors (text embeddings) and provides a new criterion for distinguishing normal and anomalous samples in unseen domains.

\subsection{Dual-Anchor Contrastive Learning}
\paragraph{Group-Gated Token Refiner.}
The \textit{Group-Gated Token Refiner (GGTR)} refines the \texttt{[CLS]} token $t_{\mathrm{cls}}$ using the normal and anomaly group tokens obtained from the hierarchical merging process, as illustrated in Figure~\ref{fig2} (b). 
In this module, the normal and anomaly group tokens act as \textit{image anchors} that provide semantic guidance for calibrating the \texttt{[CLS]} representation. 
Specifically, given a \texttt{[CLS]} token and the normal/anomaly group anchors $G_N$ and $G_A$, we compute their gating weights as follows:
\begin{align}
w_n &= \frac{t_{\mathrm{cls}}^\top G_N}{\|t_{\mathrm{cls}}\|\|G_N\|}, \quad
w_a = \frac{t_{\mathrm{cls}}^\top G_A}{\|t_{\mathrm{cls}}\|\|G_A\|}.
\end{align}

The concatenated gating weights are then normalized and passed through a linear layer followed by a sigmoid activation, after which the refined representation is obtained via a residual connection with the original \texttt{[CLS]} token:
\begin{equation}
\begin{aligned}
\hat{w} &= \mathrm{Concat}(w_n, w_a), \\
t_{\mathrm{cls}}^{\text{refined}} &= t_{\mathrm{cls}} + 
\sigma\!\left(\mathrm{Linear}\!\left(\mathrm{Norm}(\hat{w})\right)\right),
\end{aligned}
\end{equation}
where $\mathrm{Concat}$ denotes concatenation, $\mathrm{Norm}(\cdot)$ is a normalization function, and $\sigma(\cdot)$ denotes the sigmoid function. 
This refinement process allows the \texttt{[CLS]} token to adaptively incorporate information from both normal and abnormal semantic anchors, thereby enhancing its global discriminative representation for anomaly detection.

\paragraph{Dynamic State Prompt.}
Unlike conventional approaches that rely solely on static textual prompts, our method constructs \textit{dynamic state prompts} by incorporating image-dependent group tokens extracted from the input image. 
Specifically, the prompts are formulated as:
\begin{equation}
\begin{aligned}
p_n &= [V_1][V_2]\cdots[V_E][W_1][G_N][\texttt{class}], \\
p_a &= [V'_1][V'_2]\cdots[V'_E][W'_1][G_A][\texttt{class}],
\end{aligned}
\end{equation}
where $[V_i]$ and $[W_i]$ denote learnable context tokens. 
By leveraging these image-grounded state tokens, the proposed prompts dynamically adapt to the underlying visual distribution, rather than depending solely on pre-defined textual semantics. 
Such an image-conditioned formulation enhances the model’s robustness to domain shifts and significantly improves its generalization capability to open-set anomalies. Finally, the constructed prompts are fed into the text encoder to obtain the corresponding text embeddings $\mathcal{T} = \{t_n, t_a\}$ for vision-language alignment.

\subsection{Optimization}
\begin{table*}[t]
    \caption{\textbf{Image-level AUROC and F1-score, and Average Precision (AP) on both industrial and medical datasets.} The best results are marked in \textcolor{darkred}{red}, while the second-best are indicated in \textcolor{blue}{blue}. }
	\centering
	\label{tab1}
	\renewcommand{\arraystretch}{1.0}
	\resizebox{0.95\textwidth}{!}
	{
		\begin{tabular}{cccccccc>{\columncolor{blue!3}}c}
			\toprule
            &
			\multirow{2}{*}{Dataset} & 
            WinCLIP~\cite{WinCLIP} & 
            APRIL-GAN~\cite{APRIL-GAN} & 
            AnomalyCLIP~\cite{AnomalyCLIP} & 
            AdaCLIP~\cite{AdaCLIP} &
            AA-CLIP~\cite{AA-CLIP} &
            Bayes-PFL~\cite{Bayes-PFL} & 
            \textbf{Ours} \\   

            & &
            [CVPR'23] &
            [CVPR workshop'23] &
            [ICLR'24] &
            [ECCV'24] &
            [CVPR'25] &
            [CVPR'25] &
            - \\
            
            \midrule
			
            & MVTec-AD      & (91.8, 92.9, 95.1)    & (86.1, 90.4, 93.5)    & (91.5, 92.8, 96.2)    & (92.0, 92.7, 96.4)    & (90.5, 78.5, 95.9)    & (\textcolor{blue}{92.3}, \textcolor{blue}{93.1}, \textcolor{blue}{96.7})    & (\textcolor{darkred}{92.7}, \textcolor{darkred}{94.0}, \textcolor{darkred}{97.1}) \\ 
                                        
            & VisA          & (78.1, 79.0, 77.5)    & (78.0, 78.7, 81.4)    & (82.1, 80.4, 85.4)    & (83.0, 81.6, 84.9)    & (77.8, 73.9, 81.4)    & (\textcolor{blue}{87.0}, \textcolor{blue}{84.1}, \textcolor{blue}{89.2})    & (\textcolor{darkred}{88.3}, \textcolor{darkred}{85.0}, \textcolor{darkred}{91.8}) \\
			 
            & MPDD          & (61.4, 77.5, 69.9)    & (76.7, 80.9, 82.5)    & (77.0, 80.8, 82.0)    & (71.2, 77.6, 75.0)    & (55.9, 70.4, 65.4)    & (\textcolor{blue}{78.3}, \textcolor{darkred}{82.2}, \textcolor{darkred}{80.1})    & (\textcolor{darkred}{78.5}, \textcolor{blue}{81.9}, \textcolor{darkred}{84.5}) \\
            
            & BTAD          & (83.3, 81.0, 84.1)    & (74.2, 70.0, 71.7)    & (89.1, 86.0, 91.1)    & (91.6, 88.9, 92.4)    & (92.5, 55.1, 94.4)    & (\textcolor{blue}{93.2}, \textcolor{darkred}{91.9}, \textcolor{darkred}{96.5})    & (\textcolor{darkred}{95.9}, \textcolor{blue}{91.0}, \textcolor{blue}{94.6}) \\
            
            & RSDD          & (85.3, 73.5, 65.3)    & (73.1, 59.7, 50.5)    & (73.5, 59.0, 55.0)    & (89.1, 75.0, 70.8)    & (87.7, 73.6, 80.6)    & (\textcolor{darkred}{94.1}, \textcolor{darkred}{89.6}, \textcolor{blue}{92.3})    & (\textcolor{blue}{94.0}, \textcolor{blue}{89.5}, \textcolor{darkred}{94.4}) \\
			
            & KSDD2         & (93.5, 71.4, 77.9)    & (90.3, 70.0, 74.4)    & (92.1, 71.0, 77.8)    & (95.9, 84.5, 95.9)    & (96.1, 82.4, 87.6)    & (\textcolor{darkred}{97.3}, \textcolor{blue}{92.3}, \textcolor{darkred}{97.9})    & (\textcolor{blue}{97.1}, \textcolor{darkred}{92.6}, \textcolor{blue}{97.7}) \\
			                                              
            & DAGM          & (89.6, 86.4, 90.4)    & (90.4, 86.9, 90.1)    & (95.6, 93.2, 94.6)    & (96.5, 94.1, 95.7)    & (93.8, 75.7, 94.2)    & (\textcolor{blue}{97.7}, \textcolor{blue}{95.7}, \textcolor{blue}{97.0})    & (\textcolor{darkred}{98.0}, \textcolor{darkred}{96.9}, \textcolor{darkred}{97.7}) \\
			                                             
            & DTD-Synthetic & (95.0, 94.3,  97.9)   & (83.9, 89.4, 93.6)    & (94.5, 94.5, 97.7)    & (92.8, 92.2, 97.0)    & (94.3, 75.0, 98.0)    & (\textcolor{blue}{95.1}, \textcolor{blue}{95.1}, \textcolor{blue}{98.4})    & (\textcolor{darkred}{98.8}, \textcolor{darkred}{98.5}, \textcolor{darkred}{99.6}) \\ 
            
            \midrule
                                                         
            & Average       & (84.8, 82.0, 82.3)    & (82.0, 78.3, 78.8)    & (86.9, 82.1, 85.0)    & (89.0, 86.0, 88.5)    & (86.6, 73.1, 87.2)               & (\textcolor{blue}{91.2}, \textcolor{blue}{90.3}, \textcolor{blue}{92.9})    & (\textcolor{darkred}{92.9}, \textcolor{darkred}{91.1}, \textcolor{darkred}{94.6}) \\ 

            \bottomrule
		\end{tabular}
	}
\end{table*}
\begin{table*}[t]
    \caption{\textbf{Pixel-level AUROC and Average Precision (AP) on both industrial and medical datasets.} The best results are marked in \textcolor{darkred}{red}, while the second-best are indicated in \textcolor{blue}{blue}. }
	\centering
	\label{tab1}
	\renewcommand{\arraystretch}{1.0}
	\resizebox{0.9\textwidth}{!}
	{
		\begin{tabular}{cccccccc>{\columncolor{blue!3}}c}
			\toprule
			& \multirow{2}{*}{Dataset} & 
            WinCLIP~\cite{WinCLIP} & 
            APRIL-GAN~\cite{APRIL-GAN} & 
            AnomalyCLIP~\cite{AnomalyCLIP} & 
            AdaCLIP~\cite{AdaCLIP} &
            AA-CLIP~\cite{AA-CLIP} &
            Bayes-PFL~\cite{Bayes-PFL} & 
            \textbf{Ours} \\   

            & &
            [CVPR'23] &
            [VAND'23] &
            [ICLR'24] &
            [ECCV'24] &
            [CVPR'25] &
            [CVPR'25] &
            - \\
            
            \midrule

            & MVTec-AD      & (85.1, 18.0)    & (87.6, 40.8)    & (91.1, 34.5)    & (86.8, 38.1)    & (\textcolor{blue}{91.9}, 46.7)    & (91.8, \textcolor{blue}{48.3})    & (\textcolor{darkred}{92.4}, \textcolor{darkred}{50.4}) \\
			                                              
            & VisA          & (79.6,  5.0)    & (94.2, 25.7)    & (95.5, 21.3)    & (95.1, 29.2)    & (94.7, 23.8)    & (\textcolor{blue}{95.6}, \textcolor{blue}{29.8})    & (\textcolor{darkred}{96.6}, \textcolor{darkred}{34.4}) \\
			      
            & MPDD          & (71.2, 14.1)    & (91.8, 26.4)    & (\textcolor{blue}{96.5}, 23.8)    & (95.6, \textcolor{blue}{31.0})    & (95.9, 23.2)    & (\textcolor{darkred}{97.1}, 30.3)    & (95.9, \textcolor{darkred}{31.8}) \\
            
            & BTAD          & (71.4, 11.2)    & (91.3, 32.9)    & (93.3, 42.0)    & (87.7, 36.6)    & (\textcolor{blue}{94.0}, 41.6)    & (93.9, \textcolor{blue}{47.1})    & (\textcolor{darkred}{97.5}, \textcolor{darkred}{59.3}) \\
                                                        
            & RSDD          & (95.1, 2.1)     & (99.4, 30.6)    & (99.1, 19.1)    & (\textcolor{blue}{99.5}, 38.2)    & (\textcolor{blue}{99.5}, \textcolor{darkred}{40.6})    & (\textcolor{darkred}{99.6}, 39.1)    & (99.2, \textcolor{blue}{40.1}) \\
			                                          
            & KSDD2         & (97.9, 17.1)    & (97.9, 61.6)    & (99.4, 41.8)    & (96.1, 56.4)    & (98.6, 38.0)    & (\textcolor{darkred}{99.6}, \textcolor{blue}{73.7})    & (\textcolor{blue}{99.5}, \textcolor{darkred}{81.3}) \\
			                                                                                      
            & DAGM          & (83.2, 3.1)     & (\textcolor{blue}{99.2}, 42.6)    & (99.1, 29.5)    & (97.0, 44.2)    & (87.4, 40.3)    & (\textcolor{darkred}{99.3}, \textcolor{blue}{43.1})    & (98.3, \textcolor{darkred}{55.7}) \\
			                                           
            & DTD-Synthetic & (82.5, 11.6)    & (96.6, 67.3)    & (97.6, 52.4)    & (94.1, 52.8)    & (97.0, 60.6)    & (\textcolor{blue}{97.8}, \textcolor{blue}{69.9})    & (\textcolor{darkred}{98.9}, \textcolor{darkred}{82.7}) \\ 
        
            & ISIC          & (83.3, 62.4)    & (85.8, 69.8)    & (88.4, 74.4)    & (85.4, 70.6)    & (\textcolor{darkred}{94.3}, \textcolor{darkred}{87.7})     & (92.2, 84.6)    & (\textcolor{blue}{93.0}, \textcolor{blue}{84.7}) \\
			                                           
            & CVC-ColonDB   & (64.8, 14.3)    & (78.4, 23.2)    & (81.9, 31.3)    & (79.3, 26.2)    & (80.0, 20.6)     & (\textcolor{blue}{82.1}, \textcolor{darkred}{31.9})    & (\textcolor{darkred}{83.4}, \textcolor{blue}{27.4}) \\
			                                            
            & CVC-ClinicDB  & (70.7, 19.4)    & (86.0, 38.8)    & (85.9, \textcolor{blue}{42.2})    & (84.3, 36.0)    & (85.9, 41.5)     & (\textcolor{darkred}{89.6}, \textcolor{darkred}{53.2})    & (\textcolor{blue}{87.1}, 39.8) \\
			                                         
            & TN3K          & (70.7, 20.3)    & (73.6, 35.7)    & (81.5, 44.8)    & (75.4, 31.1)    & (77.9, 35.6)     & (\textcolor{blue}{78.6}, \textcolor{blue}{40.0})    & (\textcolor{darkred}{83.9}, \textcolor{darkred}{46.0}) \\
			                                          
            & Endo          & (68.2, 23.8)    & (84.1, 47.9)    & (86.3, 50.4)    & (84.0, 44.8)    & (88.2, \textcolor{darkred}{69.4})     & (\textcolor{darkred}{89.2}, 58.6)    & (\textcolor{blue}{88.8}, \textcolor{blue}{60.4}) \\
			                                           
            & Kvasir        & (69.8, 27.5)    & (80.2, 42.4)    & (81.8, 42.5)    & (79.4, 43.8)    & (83.1, 49.5)     & (\textcolor{blue}{85.4}, \textcolor{blue}{54.2})    & (\textcolor{darkred}{86.1}, \textcolor{darkred}{65.4})\\ 
            \midrule
                                                   
            & Average       & (78.1, 17.9)      & (89.0, 41.8)      & (91.2, 39.3)      & (88.6, 41.4)      & (90.6, 44.2)       & (\textcolor{blue}{92.3}, \textcolor{blue}{50.3})      & (\textcolor{darkred}{92.9}, \textcolor{darkred}{54.2}) \\ 
            
            \bottomrule
		\end{tabular}
	}
    \vspace{-4mm}
\end{table*}
\paragraph{Anomaly Map and Anomaly Score.}
Unlike conventional approaches that independently compute similarity for all patch tokens, 
our method measures the patch-level anomaly score by directly computing the similarity not only across the original patch tokens but also including the assigned token, formulated as:
\begin{equation}
\mathcal{A}^{(L+1)} \;=\; \mathrm{Softmax}\!\big(\mathrm{Up}([\frac{1}{L}\sum_{l=1}^{L}F_{i}^{(l)},\, F_{\text{assign}}]\,\mathcal{T}^{\top})\big).
\end{equation}
where $\mathrm{Up}(\cdot)$ denotes the upsampling operation to restore the spatial resolution of the anomaly map. 
The final anomaly map $\mathcal{M}$ is computed based on the normal and anomaly similarity maps as follows:
\begin{equation}
\mathcal{M} = \frac{1}{L+1}\sum_{l=1}^{L+1}\frac{\mathcal{A}^{(l)}_{a} + (1 - \mathcal{A}^{(l)}_{n})}{2},
\end{equation}
where $\mathcal{A}^{(l)}_{a}$ and $\mathcal{A}^{(l)}_{n}$ denote the anomaly and normal similarity maps, respectively.

Furthermore, to obtain the final anomaly score, we measure the similarity between the refined class token and the text embedding. 
The similarity is computed using the cosine similarity with a temperature parameter $\tau$ as follows:
\begin{equation}
p_{\text{cls}} = 
\frac{t_{\mathrm{cls}}^{\text{refined}} \cdot \mathcal{T}}
{t_{\mathrm{cls}}^{\text{refined}}\|_2 \, \|\mathcal{T}\|_2},
\end{equation} 
\paragraph{Loss function.}
We optimize the model with losses at two granularities. 
At the pixel level, we employ \textit{Focal} and \textit{Dice} losses to improve localization; 
at the image level, we use a Binary Cross-Entropy (BCE) loss for robust classification.
\begin{equation}
\begin{aligned}
\mathcal{L}_{\mathrm{cls}} &= \mathrm{BCE}(p_{\mathrm{cls}}, y), \\
\mathcal{L}_{\mathrm{seg}} &= \mathrm{Dice}(\mathcal{A}^{(L+1)}, S) + \mathrm{Focal}(\mathcal{A}^{(L+1)}, S),
\end{aligned}
\end{equation}
where $S$ denotes the ground-truth mask.

To encourage disentangled semantics between normal and abnormal group tokens, we add an orthogonality regularizer:
\begin{equation}
\mathcal{L}_{\mathrm{ortho}} \;=\; \big| \langle \mathbf{G}_{N},\, \mathbf{G}_{A} \rangle \big|^{2}.
\end{equation}

Moreover, to endow the two group tokens extracted above with explicit representations for normal and anomalous regions, we use the ground-truth segmentation mask to partition the assigned tokens into normal and anomalous subsets and compute their mean features. Let $\mathbf{Z}_N$ and $\mathbf{Z}_A$ denote the mean features of the normal and anomalous assigned tokens, respectively.
We then minimize the following cosine similarity loss:
\begin{equation}
\mathcal{L}_{\mathrm{group}} \;=\;
1 - \tfrac{1}{2}\!\left(
\cos(\mathbf{G}_N, \mathbf{Z}_N) + \cos(\mathbf{G}_A, \mathbf{Z}_A)
\right).
\end{equation}
The overall training objective is
\begin{equation}
\mathcal{L}_{\text{total}} \;=\;
\mathcal{L}_{\mathrm{cls}} + \mathcal{L}_{\mathrm{seg}} + \mathcal{L}_{\mathrm{ortho}} + \mathcal{L}_{\mathrm{group}}.
\end{equation}

\section{Experiment}
\label{sec:experiment}
\begin{figure*}[t]
\centering
\includegraphics[width=0.9\textwidth]{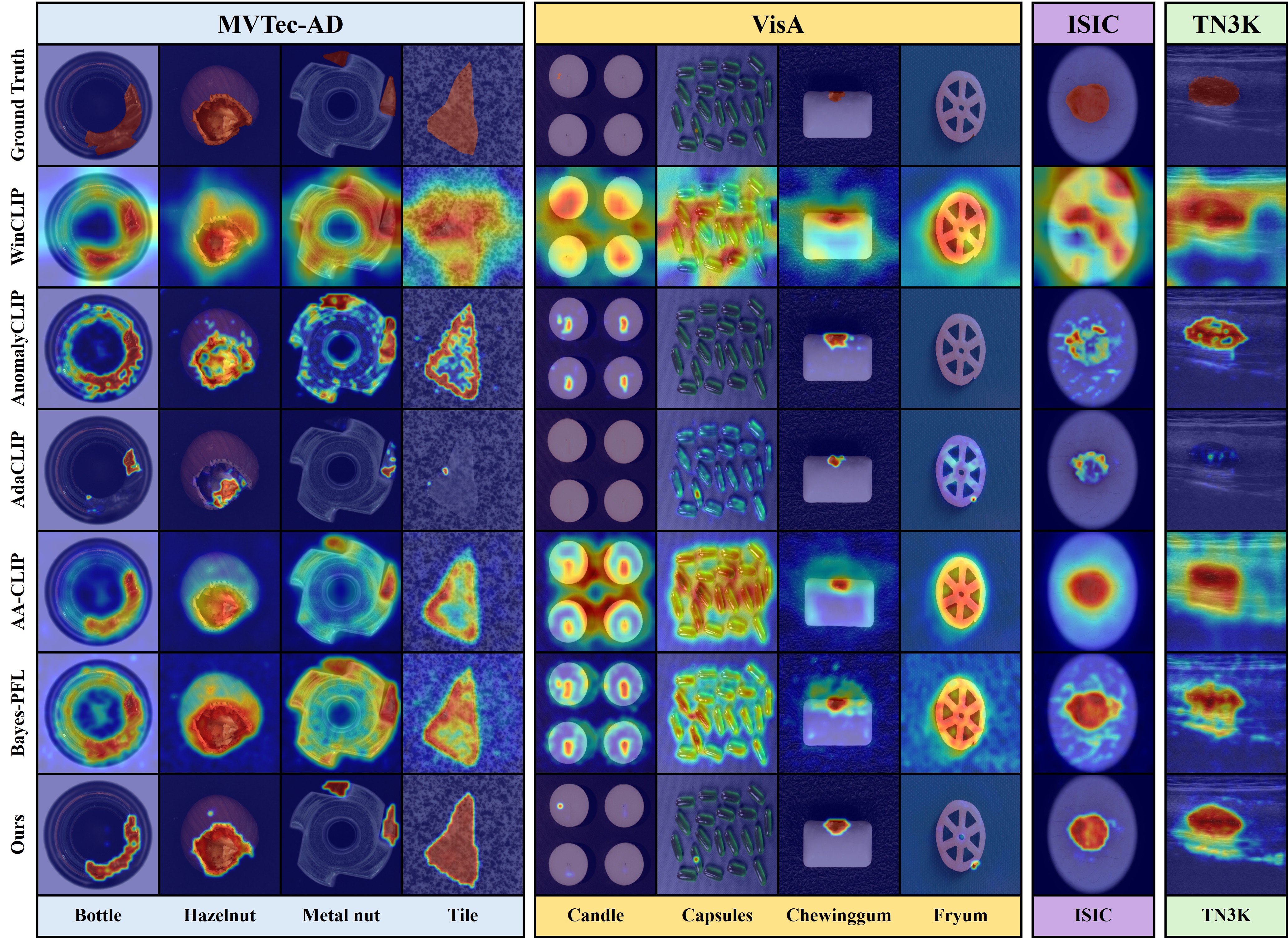}
\caption{
\textbf{
    Qualitative comparison of anomaly localization results on industrial and medical datasets.}Our method achieves more precise and consistent anomaly detection.
}
\label{fig4}
\vspace{-4mm}
\end{figure*}
\subsection{Experimental Setup}
\paragraph{Datasets.}We conduct experiments on a total of 14 datasets, comprising 8 industrial-domain datasets and 6 medical-domain datasets. 
Specifically, the industrial datasets include MVTec AD~\citep{bergmann2021mvtec}, VisA~\citep{zou2022spot}, MPDD~\citep{jezek2021deep}, BTAD~\citep{mishra2021vt}, RSDD~\citep{yu2018coarse}, KSDD2~\citep{bovzivc2021mixed}, DAGM~\citep{wieler2007weakly}, and DTD-Synthetic~\citep{aota2023zero}, while the medical datasets consist of ISIC~\citep{gutman2016skin}, ClinicDB~\citep{bernal2015wm}, ColonDB~\citep{tajbakhsh2015automated}, TN3K~\citep{gong2021multi}, Endo~\citep{hicks2021endotect}, and Kvasir~\citep{jha2019kvasir}. All datasets follow the same configurations as used in \citet{Bayes-PFL}. Detailed descriptions of each dataset are provided in the Supplementary Material A. For experiments spanning both industrial and medical domains, we employ the MVTec AD dataset as auxiliary training data, whereas for evaluation on MVTec AD itself, the VisA dataset is utilized for training.

\paragraph{Evaluation Metrics.}
For evaluation, we follow the standard zero-shot anomaly detection protocol and report AUROC and AP as the primary metric. We additionally include F1-score for image-level evaluation.
\paragraph{Implementation Details.}
We adopt a pre-trained DINOv3 (ViT-L/16) backbone to extract 1024\,-dimensional patch features from the 6\textsuperscript{th}, 12\textsuperscript{th}, 18\textsuperscript{th}, and 24\textsuperscript{th} transformer layers. 
For the text branch, we employ CLIP (ViT-L/14@336px)\footnote{\url{https://github.com/mlfoundations/open_clip}} to obtain 768\,-dimensional embeddings. All images are resized to $518 \times 518$ prior to inference. Additional implementation details including the hyperparameter settings are explained in Supplementary Material A.

\subsection{Experimental Results}
\begin{table}[t]
\caption{\textbf{Ablation on key components.}}
\centering
\resizebox{\linewidth}{!}{
\begin{tabular}{l|cc}
\toprule

\textbf{Module} &
\textbf{Image-level}& 
\textbf{Pixel-level} \\

\midrule

(a) w/o Hierarchical Group Merging  & 88.3, 90.8, 93.2 & 82.8, 44.1 \\
(b) w/o Group-Gated Token Refiner   & 88.4, 90.9, 93.5 & 80.6, 46.1 \\
(c) w/o Dynamic State Prompt        & 90.6, 92.2, 95.5  & 84.6, 44.8 \\

\midrule
\rowcolor{blue!3}
Ours & \textbf{92.7}, \textbf{94.0}, \textbf{97.1} & \textbf{92.4}, \textbf{50.4} \\

\bottomrule
\end{tabular}
}

\label{tab:ablation}
\vspace{-4mm}
\end{table}

\paragraph{Quantitative Results.} 
As shown in Table~\ref{tab1}, our model consistently outperforms all existing baselines across both industrial and medical benchmarks. In particular, the proposed \textit{Hierarchical Group Merging} mechanism demonstrates strong generalization not only on object-centric datasets such as MVTec-AD and VisA, but also on texture-oriented datasets including DAGM and DTD-Synthetic, highlighting its robustness across diverse visual domains. 
Furthermore, the superior performance achieved in medical datasets validates that our \textit{Dual-Anchor framework} effectively establishes more balanced and discriminative decision boundaries for anomalies in unseen domains, compared to conventional single-anchor approaches.

\paragraph{Qualitative Results.}As illustrated in Figure~\ref{fig4}, our qualitative results demonstrate superior localization performance compared to existing approaches. 
For instance, in the \textit{bottle} class, previous methods either fail to fully capture the anomalous regions or incorrectly assign high anomaly scores to normal areas, 
whereas our method precisely localizes only the true anomalous regions. 
This result verifies that the proposed hierarchical group merging effectively captures anomaly-aware semantic information. 
Furthermore, the results on the \textit{tile} class indicate that our approach is more robust to background noise and false positive regions. 
Such robustness validates the effectiveness of our dual-anchor strategy in overcoming the limitations of single anchor strategy. More qualitative results are provided in Supplementary Material B.

\subsection{Ablations Analysis}
\begin{table}[t]
\centering
\caption{\textbf{Ablation on the number of group tokens.}}
\resizebox{0.9\linewidth}{!}{
\begin{tabular}{c|cc|cc}
\toprule
\# Groups  & \multicolumn{2}{c|}{Image-level} & \multicolumn{2}{c}{Pixel-level} \\
{[}$G_1,G_2,G_3,G_4${]} & AUROC & AP & AUROC & AP \\
\midrule
{[}2, 2, 2, 2{]} & 88.2 & 90.8 & 83.9 & 42.7 \\
\rowcolor{blue!3}
{[}16, 8, 4, 2{]} & \textbf{92.7} & \textbf{97.1} & \textbf{92.4} & \textbf{50.4} \\
{[}128, 32, 8, 2{]} & 92.3 & 96.2 & 86.8 & 39.5 \\
\bottomrule
\end{tabular}
}
\label{tab:group}
\vspace{-4mm}
\end{table}
\paragraph{Ablation on Hierarchical Group Merging.}
To assess the contribution of the proposed Hierarchical Group Merging (HGM), we remove the assigned token produced by the top-down fusion process, as shown in Table~\ref{tab:ablation}(a). This leads to a clear performance drop, confirming that hierarchical semantic fusion is crucial for constructing discriminative anomaly representations beyond independent patch tokens.

\begin{figure}[t]
\centering
\includegraphics[width=0.9\columnwidth]{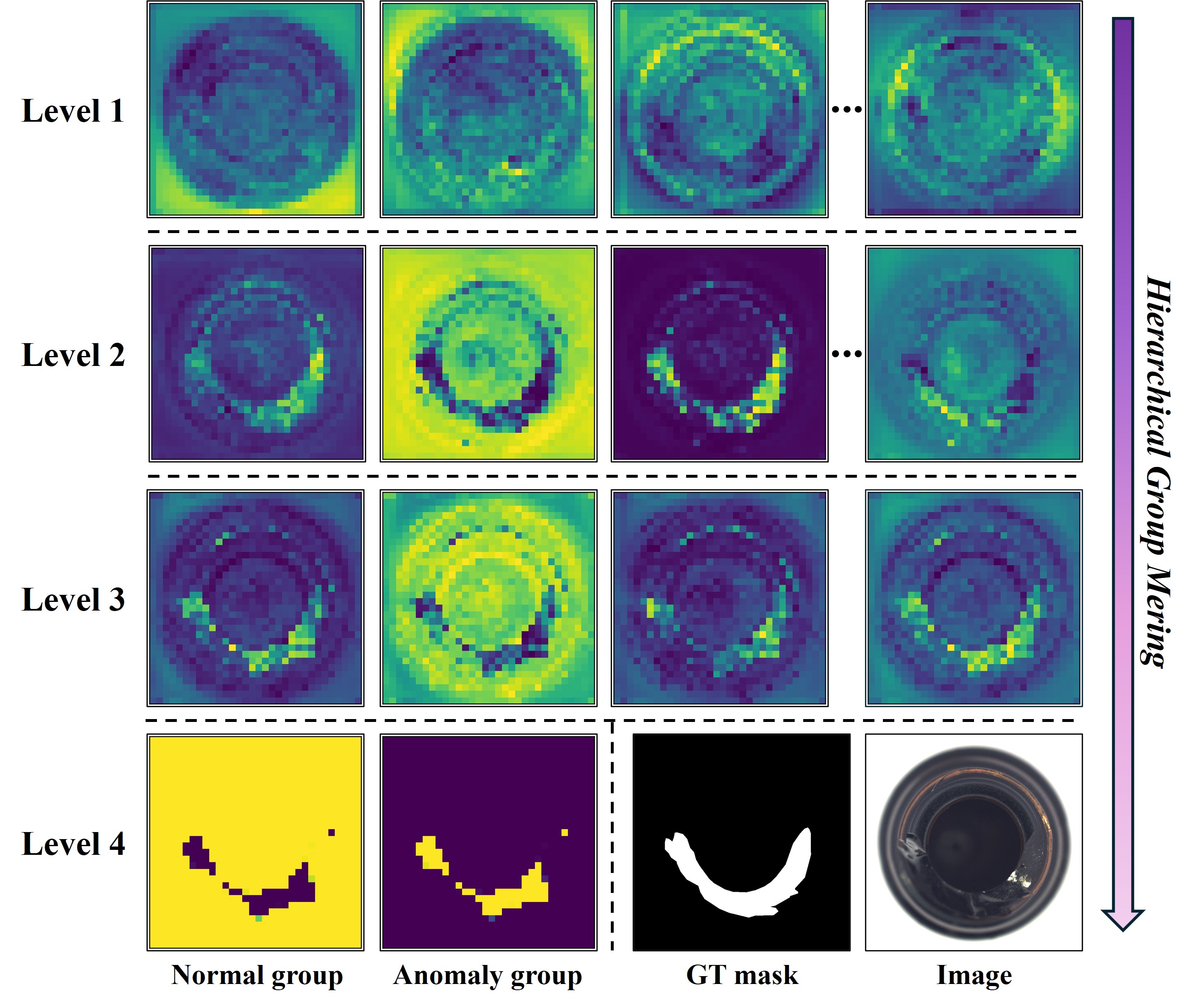}
\caption{
    \textbf{Visualization of attention maps across hierarchical levels in the Hierarchical Group Merging process.}
}
\label{fig5}
\vspace{-4mm}
\end{figure}

Figure~\ref{fig5} further visualizes the assignment attention maps across merge stages in an unseen domain. The group tokens initially capture diverse features, but progressively form normal/anomaly specific groups, with the final stage closely aligning with the ground-truth masks. These findings demonstrate that HGM effectively aggregates semantic information and enhances anomaly discrimination even under unseen conditions.

\paragraph{Ablation on Group-Gated Token Refiner.}
As shown in Table~\ref{tab:ablation}(b), removing the Group-Gated Token Refiner substantially degrades both image-level and pixel-level performance, confirming that the normal and anomaly group tokens act as effective image anchors that enhance the discriminability of the decision boundary.
Table~\ref{tab:group} further analyzes the effect of varying the number of group tokens across merge blocks. With the final stage fixed to two groups, increasing or decreasing the initial group count from an appropriate setting either weakens hierarchical fusion or introduces unnecessary redundancy. These results highlight the importance of choosing a balanced number of group tokens for stable semantic aggregation.

\paragraph{Ablation on Dynamic State Prompt.}
As shown in Table~\ref{tab:ablation}(c), removing the Dynamic State Prompt weakens the model’s ability to generate image-dependent, domain-aware textual conditioning in unseen domains. Without this adaptive mechanism, the text representation becomes static and less aligned with the input image characteristics, ultimately leading to less reliable and less consistent normal/anomaly discrimination. Additional ablation results are provided in Supplementary Material C.

\section{Conclusion}
\label{sec:conclusion}

We proposed a \textbf{Dual-Anchor framework} for ZSAD that balances image--text alignment through hierarchical image anchors. Our method constructs normal and abnormal group tokens via \textit{Hierarchical Group Merging}, refines them as visual anchors with a \textit{Group-Gated Token Refiner}, and applies a \textit{Dynamic State Prompt} for adaptive text conditioning that reduces prompt dependence and improves robustness to domain shifts. Experiments on 8 industrial and 6 medical benchmarks show consistent gains over prior ZSAD approaches.

\textbf{Limitation.} Our method, like prior ZSAD approaches, still struggles with high-level logical defects in unseen domains. As future work, we plan to extend it to a query-image-based few-shot setting to better handle such cases.

\section*{Acknowledgements}
This research was supported by the IITP(Institute of Information \& Communications Technology Planning \& Evaluation)-ITRC(Information Technology Research Center) grant funded by the Korea government(Ministry of Science and ICT) (IITP-2026-RS-2023-00260091, 34\%), the National Research Foundation of Korea(NRF) grant funded by the Korea government(MSIT)(RS-2025-16066849, 33\%), and the Institute of Information \& Communications Technology Planning \& Evaluation(IITP) grant funded by the Korea government(MSIT) (No.RS-2025-02263706, Development of an analog-digital mixed ultra-low power neuromorphic edge SoC, 33\%).

{
    \small
    \bibliographystyle{ieeenat_fullname}
    \bibliography{main}
}
\newpage

\end{document}